\def\year{2021}\relax
\documentclass[letterpaper]{article} 
\usepackage{aaai21}  
\usepackage{times}  
\usepackage{helvet} 
\usepackage{courier}  
\usepackage[hyphens]{url}  
\usepackage{graphicx} 
\usepackage{natbib}  
\usepackage{caption} 
\usepackage[switch]{lineno}
\usepackage{xcolor}
\usepackage{amssymb}
\usepackage{amsmath}
\usepackage{dsfont}
\usepackage{latexsym}
\usepackage{graphicx}
\usepackage{accents}
\usepackage{amsthm}
\newtheorem{theorem}{Theorem}[section]

\title{The Gap on \textsc{Gap}: Tackling the Problem of Differing Data Distributions \\ in Bias-Measuring Datasets}
\author{
    Vid Kocijan,\textsuperscript{1} \ \ Oana-Maria Camburu,\textsuperscript{1,2} \ \ Thomas Lukasiewicz\textsuperscript{1,2}
    \\
}
\affiliations{
  \textsuperscript{1}University of Oxford, UK \\ 
  \textsuperscript{2}Alan Turing Institute, London, UK \\
  \texttt{firstname.lastname@cs.ox.ac.uk}

}

\begin{document}
\maketitle

\begin{abstract}
Diagnostic datasets that can detect biased models are an important prerequisite for bias reduction within natural language processing.
However, undesired patterns in the collect\-ed data can make such tests incorrect.
For example, if the feminine subset of a gender-bias-measuring coreference resolution dataset contains sentences with a longer average distance between the pronoun and the correct candidate, an RNN-based model may perform worse on this subset due to long-term dependencies.
In this work, we introduce a theoretically grounded method for weighting test samples to cope with such patterns in the test data.
We demonstrate the method on the \textsc{Gap} dataset for coreference resolution. 
We annotate \textsc{Gap} with spans of all personal names and show that examples in the female subset contain more personal names and a longer distance between pronouns and their referents, potentially affecting the bias score in an undesired way.
Using our weighting method, we find the set of weights on the test instances that should be used for coping with these correlations,  
and we re-evaluate $16$ recently released coreference models.\footnote{The annotations, the weights, and the code can be found at \url{https://github.com/vid-koci/weightingGAP}.}
\end{abstract}
\section{Introduction}

AI systems trained on biased or imbalanced data can propagate and amplify observed patterns and make biased decisions at the time of evaluation and deployment~\cite{BiasSurvey}.
To detect the underlying bias in released natural language processing (NLP) systems and to increase their fairness, several diagnostic datasets have been introduced, commonly focusing on gender bias~\cite{WinoGender, WinoBias, Gap, SemanticAnalysisBias}.
In addition to overall performance, these works also define the \textit{bias score} of the evaluated model, usually the difference or ratio between the performance of the model on the groups of interest, e.g., female and male subsets of the data. However, when the subsets corresponding to the groups of interest consist of data coming from different distributions, undesired imbalances may appear, leading to inaccurate bias scores.
For example, \citet{Gap} constructed the \textsc{Gap} coreference dataset by collecting examples from English Wikipedia and observe that a random baseline does not achieve a balanced bias score, despite being unbiased by design.
They note that feminine sentences in the dataset contain more personal names and therefore more distractor mentions.
If a coreference model is negatively affected by a larger number of potential candidates, it could appear more biased against female examples than it actually is. In this work, we introduce a method for coping with imbalances in such bias-detection datasets.
We tailor the demonstration around gender bias, however, the method can be applied to any other type of bias as well. 

A possible solution to the problem of imbalances caused by different data distributions could be the augmentation of the data by introducing examples with swapped genders.
While this method has been applied to training data~\cite{WinoBias}, for test data, it could have an unforeseen impact on different NLP systems.
For example, such data augmentation may introduce instances that contain biologically or historically inaccurate facts, such as ``men giving birth'' or ``historical figures being of the opposite gender''.
Since \textsc{Gap} was collected from Wikipedia, a large number of examples with swapped gender could suffer from this problem.

Alternatively, examples can be constructed from manually crafted templates where genders can be swapped without risking to introduce inaccurate facts~\cite{WinoGender,WinoBias,SemanticAnalysisBias}.
While such tests of bias are important due to their controlled environment, they are synthetic and hence may not give a full picture of the underlying biases that can emerge when the model is used in practice.

The method introduced in this work assigns weights to test samples to cope with undesired imbalances in bias-mea\-sur\-ing datasets. 
Given a list of properties that should not correlate with the gender of data examples, we derive a set of linear equations that should hold for weights of the test samples.
At the same time, we minimize the likelihood of introducing noise into the measured bias of the models, by deriving an optimization objective that minimizes the upper bound of such noise.
The search for optimal weights under given constraints is then implemented as a linear program.

We demonstrate the use of this approach on the \textsc{Gap} dataset for coreference resolution~\cite{Gap}, the currently largest dataset for gender-bias-measuring.
We annotate the \textsc{Gap} test set with all mentions of personal names. 
The annotations will be publicly released and may potentially be used for other research directions, such as the evaluation of named-entity recognition (NER) systems.

We show that, in the \textsc{Gap} test set, the feminine examples contain, on average, $0.75$ more names per sentence than the masculine ones. 
Similarly, the correct candidate usually stands $0.5$ candidates further away from the pronoun in feminine examples than in masculine ones. 
These are all imbalances that can affect the score of a model. 
We show the effectiveness of our weighting method by showing that a series of unbiased baselines indeed achieve scores closer to the unbiased score on the weighted test set.
Finally, we re-evaluate $16$ recently released coreference models on the weighted test set of \textsc{Gap}.
We observe that several models change their bias scores when evaluated on the weighted test set, although most of these changes are small.
We encourage future research to use the introduced weighted bias metric instead.

The contributions of this paper are briefly the following:
\begin{itemize}
    \item We introduce a novel method to reduce the harm of differing data distributions in bias-measuring datasets.
    \item We manually annotate the \textsc{Gap} test set with all mentions of personal names.
    \item We identify two properties that are imbalanced in the \textsc{Gap} test set and compute weights for this set according to the introduced method.
    \item We re-evaluate $16$ recently released models for coreference resolution with the newly introduced weights.
\end{itemize}

The rest of this paper is organized as follows. In Section~\ref{theory}, we introduce and theoretically justify the method for balancing the test set. We demonstrate the use of the method on the \textsc{Gap} dataset 
in Section~\ref{Experiments}. We discuss related work  and summarize the main results in Sections~\ref{related-work} and~\ref{Summary}, respectively.

\section{Weighting Method}
\label{theory}

In this section, we present our weighting method. 
While, for ease of presentation, we describe our method on gender-bias detection, the method can be applied to any type of bias detection, and it also generalizes to tasks where one needs to detect biases among $n \,{>}\, 2$ classes, e.g.,  racial bias, by observing every pair of classes separately.
The current version of the method assumes that accuracy is used as a metric of performance. 
We leave the analysis of other potential metrics to future work.

\subsection{Definitions and Objectives}

Let $D$ be a bias-testing dataset with $n$ examples $D\,{=}\,\{x_1,$ $\ldots,x_n\}$.
Let $A$ and $B$ be non-overlapping subsets of $D$. 
We assume that $A\cup B = D$, i.e., we ignore examples outside of observed sets, if any.
The aim is to compare the performance of a model on $A$ and $B$, in order to see if the model is biased.

Let $S_1, \ldots, S_m$ be subsets of $D$, such that $S_j$ consists of all examples with a property that is not specific to the sets $A$ and $B$ but that could have an impact on the performance of an evaluated model.
For example, in the context of coreference resolution, one of the observed properties can be the number of referents in an example.
In such an example, one set $S_k$ would consist of all examples with exactly $k$ potential referents.
Note that these sets may overlap, as properties do not have to be mutually exclusive.
We assume that these properties and sets are explicitly identified beforehand, and we refer to them as identified properties.

Let $C\,{\subseteq}\, D$ be a set of examples that a model solves correctly.
Generally, the accuracy of a model corresponds to ${\lvert C\rvert}\,/\,{\lvert D\rvert}$.
However, the performance of a model and hence $C$ may have a significantly different overlap with $S_j$ than with  $D\,{\setminus}\, S_j$.
Less formally, a model may be more/less likely to solve examples in the set $S_j$.
To obtain an accurate bias measure, properties that do not influence the bias should be evenly distributed across $A$ and~$B$.
If this is not the case for a bias-detection dataset, we adapt the bias metric so that $S_j \,{\cap}\, A$ carries equal weight as $S_j \,{\cap}\, B$ in the final score.


To achieve this, we assign to each example $x_i$ its weight $w_i$ and replace the accuracy with the weighted accuracy.
We aim to find a set of weights $W$, such that $\sum_{x_i \in A\cap S_j}w_i = \sum_{x_i\in B\cap S_j}w_i$.
Additionally, we impose the following restrictions on the weights:
\begin{itemize}
    \item Balance between the observed sets:  $$\mbox{$\sum_{x_i \in A} w_i = \sum_{x_i \in B} w_i\,$.}$$
    \item Fixed sum: $$\mbox{$\sum_{i=1}^n w_i = n\,$.}$$
    \item Non-negativity: $$w_i \geq 0\mbox{, for all }i\in \{1,\ldots, n\}.$$ 
\end{itemize}
The first two 
will simplify the future derivations, while the last one  
is put in place to avoid the situation where an incorrect answer is preferred over the correct one.
A direct consequence of the first two is that the sum of all weights of one gender is fixed to $\sum_{x_i \in A} w_i = \sum_{x_i \in B} w_i = \frac{n}{2}$.

%
%
%

There could exist several sets of weights that meet the above criteria.
Among them, we prefer the distribution that minimizes the potential exacerbation of other patterns in the data, that is, any changes in the bias score of a model that are not directly related to the above-identified properties.
Let $\text{Acc}^D(C)$ and $\text{Acc}_W^D(C)$ be the unweighted and weighted accuracy, respectively, obtained by a set of correct answers~$C$ on a set $D$.
Since bias scores compare the performance on both $A$ and $B$, we aim to minimize both 
\begin{align}
\label{equation_masculine_noise}
\lvert \text{Acc}_W^A(C \cap A) - \text{Acc}^A(C \cap A)\rvert\mbox{ and}\\[0.5ex]
\label{equation_feminine_noise}
\lvert \text{Acc}_W^B(C \cap B) - \text{Acc}^B(C \cap B)\rvert\phantom{\mbox{ and}}
\end{align}

\smallskip\noindent
for any 
$C\subseteq D$.
This objective covers two cases:
\begin{itemize}
    \item When $C$ corresponds to correct answers of a model, we minimize the difference in weighted and unweighted accuracy on the sets~$A$ and $B$.
    \item When $C$ is a set of examples with some property other than the ones captured by 
    $S_1,\ldots, S_m$, we aim to retain its original overlap with $A$ and $B$.
    The overlap between sets with unidentified properties and the sets $A$ and $B$ should not be removed, as they may be an important indicator of the underlying bias. 
    An example of such a property in the context of gender bias in NLP is the amount of out-of-vocabulary words, which could be larger for feminine examples, should the text about men be more prevalent in the data used to construct the vocabulary.
\end{itemize}

Our method minimizes the upper bound on the differences~\eqref{equation_masculine_noise} and~\eqref{equation_feminine_noise}.
By considering the upper bound rather than the average case, we avoid making assumptions about the distribution of $C$.

\bigskip
\begin{theorem}[Upper Bound on Introduced Noise]\label{upper_bound_theorem}
To minimize the upper bounds of $$ \lvert \text{Acc}_W^A(C \cap A) - \text{Acc}^A(C \cap A)\rvert\mbox{ and}$$ $$ \lvert \text{Acc}_W^B(C \cap B) - \text{Acc}^B(C \cap B)\rvert\phantom{\mbox{ and}}$$ 

\smallskip \noindent for any unknown set $C\subseteq D$, it is sufficient to minimize
\begin{align*}
    \sum_{\substack{x_i,x_j\in A\\i>j}}\text{max}(w_i,w_j)+\sum_{\substack{x_i,x_j\in B\\i>j}}\text{max}(w_i,w_j).
\end{align*}
\end{theorem}

Below, we provide an intuition behind the proof.
The full proof 
is given in Appendix~\ref{upper_bound_theorem_proof}.

First, we notice that the objectives~\eqref{equation_masculine_noise} and~\eqref{equation_feminine_noise} are independent, and we can focus on each one separately.
Let us illustrate the rest of the derivation on the  objective~\eqref{equation_masculine_noise}. 
If we combine the definitions and simplify this objective, it is sufficient to minimize
\begin{align*}
    \lvert \sum_{x_i \in C\cap A}(w_i - \lambda)\rvert;\hspace{1em} \lambda=\frac{n}{2|A|}, 
\end{align*}
where $\lambda$ is constant.
To minimize the upper bound on this objective, we look at the worst-case scenario.
For $A\cap C$ of size $k$, there are two possible worst-case scenarios.
Either $A\cap C$ consists of examples with $k$ largest weights, or it consists of examples with $k$ smallest weights.
Since the sum of weights for each gender is fixed, maximizing smallest $k$ weights for every $k\in \{1,\ldots, |A|\}$ is equivalent to minimizing the largest $k$ weights for every $k\in\{1,\ldots, |A|\}$.
When combined for all possible values of $k$, the objective 
(to minimize) is equivalent to $\sum_{i=1}^{|A|}(w_{A(i)}-\lambda)\cdot i$, where $w_{A(i)}$ is the $i$-th smallest weight corresponding to an example in $A$.
Finally, we show that this is equivalent to minimizing
\begin{align*}
    \sum_{{x_i,x_j\in A,\,i>j}}\text{max}(w_i,w_j)-\lambda \frac{|A|(|A|-1)}{2}\,.
\end{align*}
The second part of the term is constant and can be ignored.
The final objective function is obtained by summing the objectives for the sets $A$ and $B$.

\subsection{Solving the Optimization Problem}

All listed conditions and criteria can be phrased as a linear program.
Balance between the subsets, fixed sum, non-negativity, and removing correlations are linear constraints.
The optimization objective can be phrased as a linear function by introducing auxiliary variables $m_{i,j}; 1\leq i,j \leq n$ for $\text{max}(w_i,w_j)$.
The following constraints have to hold for each of them: $m_{i,j}\geq w_i$ and $m_{i,j}\geq w_j$.

To summarize, we collect all derived constraints for the linear program:
\begin{itemize}
    \item $\sum_{x_i \in A} w_i = \sum_{x_i \in B} w_i$.
    \item $\sum_{i=1}^n w_i = n$.
    \item $w_i \geq 0$ for all $i\in \{1,\ldots, n\}$.
    \item $\sum_{x_i\in A\cap S_j}w_i = \sum_{x_i\in B\cap S_j}w_i$ for all $S_j$. 
    \item For all $i,j$, such that $i<j$ and either $w_i,w_j\in A$ or $w_i,w_j\in B$: $m_{i,j}\geq w_i$ and $m_{i,j}\geq w_j$.
\end{itemize}
The criterion function is equal to
\begin{align*}
    \text{min}\sum_{\substack{x_i,x_j\in A\\i>j}}m_{i,j}+\sum_{\substack{x_i,x_j\in B\\i>j}}m_{i,j}\,.
\end{align*}
A linear-program solver can then be used to find the minimum to this function.

\section{Experiments}
\label{Experiments}
In this section, we 
demonstrate the use of our weighting method on the \textsc{Gap} dataset~\cite{Gap}.
First, we show that feminine examples contain more candidates than masculine examples, and that the correct candidate usually stands further away from the pronoun in feminine examples than in masculine ones.
We show that weighting solves these imbalances, as several unbiased baselines obtain scores closer to $1$ after weighting ($1$ is a balanced score).
Finally, we re-evaluate 16 publicly released models for coreference resolution, observing that the majority of these models were only slightly affected by these properties.

\subsection{The \textsc{Gap} Dataset}

\textsc{Gap} is a corpus of challenging examples of pronouns from English Wikipedia. It was introduced as a gender-balanced dataset, so that exactly half of the pronouns are masculine, and half are feminine \cite{Gap}.
The test set, which the rest of the paper is about, consists of $2000$ text spans.
The dataset comes with a development and validation set; however, they are not the focus of this work.
For each text span, one pronoun has to be resolved.
Pronoun resolution is treated as a binary classification task, with the goal to determine whether a single candidate is the referent of the pronoun or not.
Note that candidates are not given as input and the model is expected to find them on its own.
It is guaranteed that candidates are always personal names from the input text and that at most one of them is the correct referent.
\citet{Gap} define a bias measure as ratio between the $F_1$-scores on the feminine and masculine subsets, ${F_1^F}\,/\,{F_1^M}$.
An unbiased system is therefore expected to achieve a bias score around~$1$.
An example from \textsc{Gap} can be found~below:

\smallskip
\textit{Kathleen first appears when Theresa and Myra visit \textbf{her} in a prison.}

\textit{Kathleen: \textbf{True}, Theresa: \textbf{False}}
\smallskip

During the scoring, the output of any evaluated model is compared to two candidates, specified by the example.

Note that any incorrect candidate adds noise to the bias score.
If a model answers \textit{Theresa}, it will be penalized with an additional false-positive outcome, unlike a model that answered \textit{Myra}, despite both being equally wrong.
Since there is never more than one correct candidate per sentence, and the candidates are not known in advance, comparing the prediction only with the correct candidate is thus not just sufficient, but also a more accurate bias measure.
So, for measuring bias, we replace the $F_1$ score with accuracy, which has already been used as a performance metric in coreference resolution before~\cite{knowref,DPR,WinoGrande}.

To be able to observe the effect of our weighting method, we first introduce a \textit{plain} accuracy-based bias metric \textit{acc-Bias}.
We measure the accuracy on positive candidates in the masculine subset $A_M$ and the accuracy on positive candidates in the feminine subset~$A_F$ and define acc-Bias as ${A_F}\,/\,{A_M}$.
Results of this metric will be compared to a later-introduced weighted accuracy.
Text spans with no positive examples are dropped, reducing the size of the test set by approximately $10\%$.

A possible improvement of \textsc{Gap} that we do not address in this work is a more fine-grained analysis and stricter definition from a linguistic perspective.
The motivation by~\citet{Gap} is focused on biosocial gender, that is, comparing performance of models when the candidates are masculine and when they are feminine.
However, in practice, the dataset measures the impact of grammatical gender, as the author define the gender of an example to match the gender of the pronoun in question.
While these two types of gender largely overlap in English, mismatch can happen, e.g., in the case of \emph{personalization} or \emph{misgendering}.
We refer to \citep{LinguisticGender} for a detailed definition, comparison of these types of gender, and the analysis of the mismatches.

We found examples containing such mismatch not to have a strong presence in the \textsc{Gap} test set, however, the exact number is hard to estimate due to the lack of context in many of the examples.
We highlight that addressing this is necessary before the dataset-creating approach by~\citet{Gap} can be used on languages with a stronger presence of grammatical gender, e.g., Russian and German.

\subsection{Baselines}

We re-implement the random and token distance baselines introduced by \citet{Gap}.
First, we find all personal names in the input text using an off-the-shelf named entity recognition (NER).
Each baseline is implemented with two NER systems: Google Cloud NL API\footnote{\url{https://cloud.google.com/natural-language/}} and Spacy en\_core\_web\_lg\footnote{\url{https://spacy.io/}}, abbreviated Spacy-lg.
Additionally, as we have manually labeled all spans in the \textsc{Gap} test set that correspond to a personal name, we use
these annotations to implement \textit{Ground-Truth} baselines, which are thus not affected by potential mistakes of the NER systems.

In the random baseline implementation, a random personal name is picked from the list.
Note that our implementation of the random baselines exhibits a different performance than the one from \citet{Gap}.
They report adding heuristics to eliminate obviously incorrect candidates; we do not follow them to avoid adding any noise.

In the token distance baseline implementation, the personal name closest to the pronoun is selected.
Distance is measured in number of tokens, using the Spacy tokenizer.
We rename this baseline as Dist-1 baseline and introduce Dist-2 and Dist-3 baselines, where we pick the second closest and third closest personal name, respectively.
If there are fewer than $2$ or $3$ candidates in the sentence, then we consider all answers to be \texttt{False}, that is, we give no answer.
We do not introduce higher-order distance-based baselines.
Their accuracy drops and with it the denominator in the bias score.
This amplifies the noise caused by mistakes of the NER system and makes their results inconclusive.

Assuming unbiased NER systems and balanced data, the baselines should achieve a bias score very close to~$1$.
The results of all baselines on the \textsc{Gap} test set is reported in the left part of Table~\ref{table-baselines-reweighted}, where
we see that most of the bias scores strongly differ from $1$.
In the next section, we show that imbalanced data are the reason behind this.
Notice that the acc-Bias score of a model is usually further from $1$ than its $F_1$-Bias score.
These results empirically support our intuition that $F_1$-Bias is less representative than acc-Bias, as noise from negative candidates makes $F_1$-Bias less sensitive.
Thus, the accuracy-based bias metric is more appropriate than its $F_1$-score counterpart.

\begin{table*}[ht!]
\centering
    \begin{tabular}{@{}l@{\ }|@{\ }c@{\ \ }c@{\ \ }c@{\ \ }c@{\ }|@{\ }c@{\ \ }c@{\ \ }c@{\ }|@{\ }c@{}}
    Baseline & $F_1$ & Accuracy & $F_1$-Bias & acc-Bias & W-Bias & W\textsubscript{num}-Bias & W\textsubscript{dist}-Bias & $W_t$-Bias\\ \hline
    Ground-Truth Random & $0.305$ & $0.224$ & $0.884$ & $0.849$ & $\mathit{1.000}$ &$\mathit{0.995}$& $0.899$ & $\mathit{1.000}$ \\ \hline
    Spacy-lg Random & $0.286$ & $0.211$ & $0.904$ & $0.870$ & $\mathit{0.975}$ & $\mathit{0.980}$ & $0.905$ & $\mathit{0.984}$\\
    Google-NER Random & $0.295$ & $0.218$ & $0.937$ & $0.907$ & $\mathit{1.019}$ &$\mathit{1.021}$& $0.949$ & $\mathit{1.020}$ \\ \hline
    Ground-Truth Dist-1 & $0.463$ & $0.412$ & $0.850$ & $0.776$ & $\mathit{1.000}$&$0.804$&$\mathit{1.000}$ &$\mathit{1.000}$\\ \hline
    Spacy-lg Dist-1 & $0.423$ &$0.375$ & $0.887$ & $0.816$ & $\mathit{1.015}$ & $0.824$&$\mathit{1.029}$ & $\mathit{1.018}$\\
    Google-NER Dist-1 & $0.446$ & $0.399$ & $0.875$ & $0.799$ & $\mathit{0.986}$&$0.793$&$\mathit{1.016}$ &$\mathit{0.994}$\\ \hline
    Ground-Truth Dist-2 &$0.353$ &$0.310$ & $0.923$ & $0.882$ & $\mathit{1.000}$ & $0.920$ & $\mathit{1.000}$ & $\mathit{1.000}$ \\ \hline
    Spacy-lg Dist-2 &$0.319$&$0.263$& $0.917$ & $0.907$ & $\mathit{0.962}$ & $0.932$&$\mathit{0.977}$&$\mathit{0.968}$\\
    Google-NER Dist-2 &$0.354$ &$0.309$ & $0.946$ & $0.915$ & $\mathit{1.000}$ & $0.983$ & $\mathit{1.001}$ & $\mathit{1.026}$ \\ \hline
    Ground-Truth Dist-3 & $0.228$ &$0.156$ & $1.270$ & $1.347$ & $\mathit{1.006}$ & $1.266$ & $\mathit{1.010}$ & $\mathit{1.007}$\\ \hline
    Spacy-lg Dist-3 &$0.205$ & $0.134$ & $1.490$ & $1.585$ & $\mathit{1.118}$ & $1.494$ & $\mathit{1.152}$ & $\mathit{1.200}$\\
    Google-NER Dist-3 & $0.219$ &$0.150$ & $1.312$ & $1.426$ & $\mathit{1.154}$ & $1.368$ & $\mathit{1.111}$ & $\mathit{1.116}$\\ \hline
    \end{tabular}
    \caption{Performance and bias metrics on baseline systems on \textsc{Gap}, implemented with two different NER systems as well as the ground-truth personal names. The reported performance of the random classifier is obtained by averaging the performance over $10,000$ repetitions.
    If the evaluated baseline is expected to achieve a score of $1$ on some metric due to balancing, the score is written in \textit{italics}.
    Note that deviations can happen when NER-system extractions are incorrect.
    }
    \label{table-baselines-reweighted}
\end{table*}

\subsection{Analysis of \textsc{Gap}}\label{Analysis}
Our analysis of the manually annotated spans of personal names shows that masculine examples contain $5.55$ personal names on average (standard deviation $3.18$), while feminine examples contain $6.30$ names on average (standard deviation $3.44$).
This confirms the hypothesis about imbalances in the data and  explains why the Ground-Truth random baseline achieved a bias score different from $1$.
The full distribution of the number of names per sentence is given in Appendix~\ref{Appendix-distribution}.

Secondly, we sort all annotated personal names in each sentence by distance to the pronoun in the same way as done by the Dist-$k$ baselines.
We find the position of the correct candidate on this ordered list.
The average position of the correct candidate in the masculine subset is $1.86$ (standard deviation $1.19$) candidates away from the pronoun, while the average position in the feminine subset is $2.32$ (standard deviation $1.54$) candidates away from the pronoun, potentially explaining the bias scores of the Dist-k baselines.
Examples with no correct candidate are not considered in this statistic.
The full distribution is given in Appendix~\ref{Appendix-distribution}.

\subsection{Weighting \textsc{Gap}}
\label{section-weighting}
Using our manual annotations of personal names, let $N_k$ be the set of all examples with exactly $k$ personal names, and let $D_k$ be the set of all examples where the correct candidate is the $k$-th closest candidate to the pronoun.
The~$N_k$ and $D_k$ sets form the sets that we generically denoted as $S_1, \ldots, S_m$ in Section~\ref{theory}. Thus, the sets $N_k$ and $D_k$ are used as input to our balancing method to obtain a linear program, which we solve with the \textsc{linprog} optimization tool from Matlab, version R2019b.

We note that balancing of \textsc{Gap} with downsampling does not scale.
To obtain a dataset that is balanced only w.r.t.\ the number of candidates in a sentence or only w.r.t.\ distance, the dataset has to be downsampled to 75\% of its size. To obtain a dataset that is balanced w.r.t.\ both, the dataset has to be downsampled below 70\% of its size, which is a significant drop in the number of samples. Should one want to remove an additional undesired pattern, this number would likely drop even further, making the pruning method unscalable.

We name the obtained weighted bias metric \textit{W-Bias}.
We highlight that the introduced constraints are not a guarantee that W-Bias is completely balanced, as other imbalances in the data may exist.
However, given Theorem~\ref{upper_bound_theorem}, known imbalances have been balanced out, while introducing the least noise possible, making the introduced metric preferred over the existing one, i.e., no weighting.
A~visualization of the weights is in Section~\ref{weights_appendix}.

To assess the introduced weights, we evaluate the baselines on the newly introduced W-Bias metric.
To confirm that our method does not introduce noise relative to  unidentified properties, we perform two ablation experiments. In the first one, 
we ignore the distance property, while in the second experiment, we ignore the number of candidates. 
To this end, we introduce two more bias metrics: W\textsubscript{num}-Bias and W\textsubscript{dist}-Bias.
In W\textsubscript{num}-Bias, the sets $D_k$, $k\in \mathbb{N}$, were not included as the input to the balancing procedure.
W\textsubscript{num}-Bias is only balanced with respect to the number of names per sentence.
On the other hand, W\textsubscript{dist}-Bias does not include the sets $N_k$, $k\in \mathbb{N}$, meaning that it is only balanced with respect to the distance between the pronoun and the correct answer.
We show that, for random baselines, the following holds: $|1-\text{W-Bias}|\leq |1-\text{W\textsubscript{dist}-Bias}|\leq |1-\text{acc-Bias}|$, that is, balancing relative to distance does not exacerbate bias scores of random baselines, and that additional balancing relative to number of names further decreases its distance to unbiased score (of $1$).
Similarly, we show that for Dist-$k$ baselines, 
$|1-\text{W-Bias}|\leq |1-\text{W\textsubscript{num}-Bias}|\leq |1-\text{acc-Bias}|$.

The results are reported in Table~\ref{table-baselines-reweighted}.
In the columns that correspond to W\textsubscript{num}-Bias and W\textsubscript{dist}-Bias, numbers in italics are expected to be similar to the numbers predicted by W-Bias.
We see that the inequations in the previous paragraph hold for all baselines, showing that our weights indeed do not exacerbate the bias of unidentified properties.
Moreover, we see that the W-Bias scores achieved by the baselines are consistently closer to $1$ than their acc-Bias scores, confirming that the introduced weights balance the bias metric.
In particular, the W-Bias score of Ground-Truth baselines is equal to $1$, i.e., unbiased.
We note that the minimal deviation from $1$ of the Ground-Truth W-bias score for the Dist-3 baseline is a consequence of a disagreement between our span annotations with the spans of gold labels.
Bias scores of the Dist-2 and Dist-3 baselines implemented with NER systems are subject to larger deviations that happen, because these baselines are more sensitive to disagreement between the NER system and our annotations of the name spans.

We note that weighting with respect to one of the imbalances sometimes helped balancing the baseline that was affected by the other.
For example, balancing the number of names per sentence (W\textsubscript{num}-Bias) resulted in improved bias scores of all Ground-Truth Distance baselines. 
This implies that there exists a correlation between the number of personal names in the sentence and the distance between the pronoun and the correct candidate in the \textsc{Gap} test set.

\subsection{Analysis of Weights}
\label{weights_appendix}

\begin{figure}[ht]
  \centering
  \includegraphics[width=0.45\textwidth]{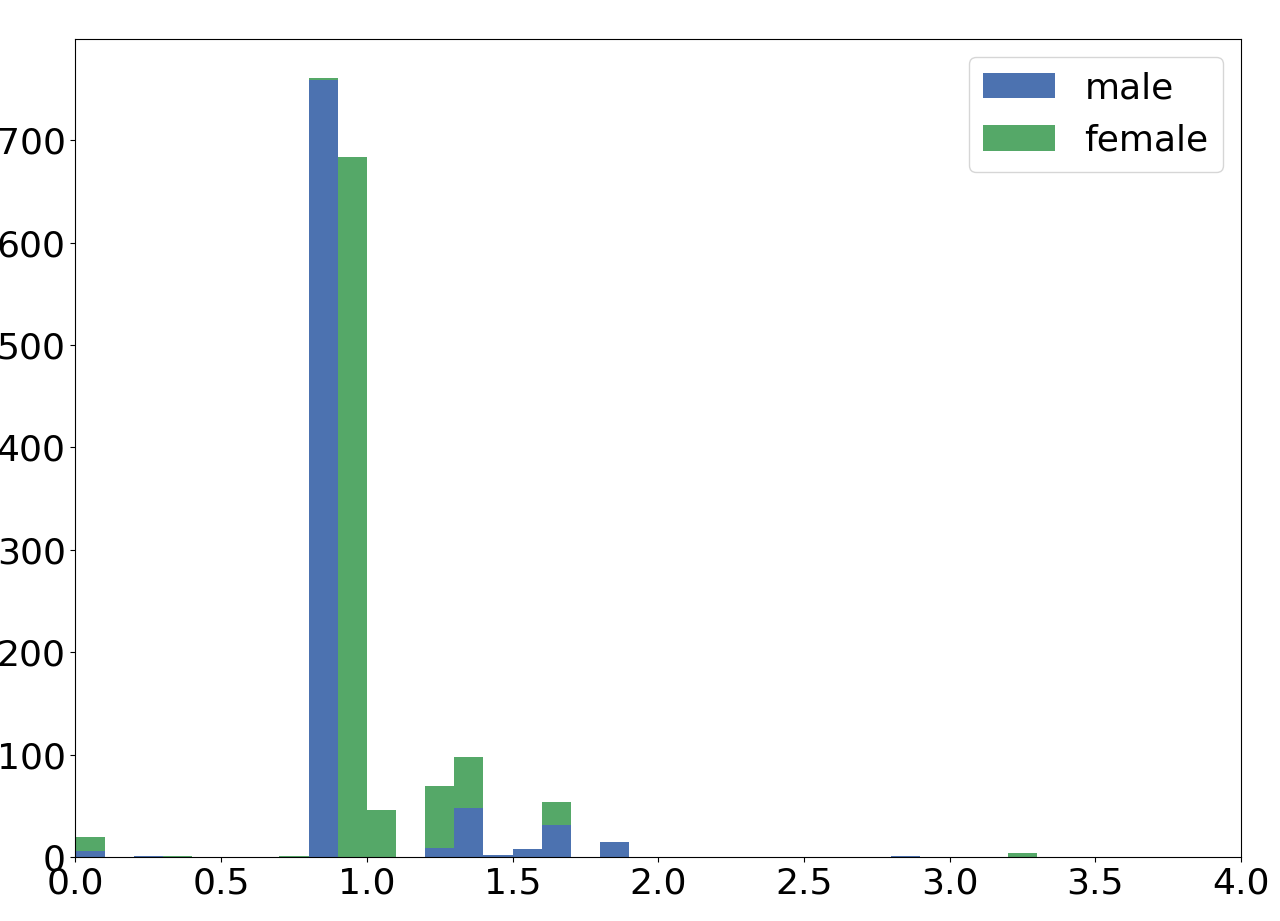}
  \caption{Histogram showing the distribution of W-Bias weights, split into intervals of size $0.1$. 
  The blue parts of the columns correspond to masculine examples, while the green parts correspond to feminine examples. 
  The weights are centered around $1$. Nine largest weights are not included in the histogram, as they have values over $4$.}
  \label{weights}
\end{figure}

\begin{figure}
  \centering
  \includegraphics[width=0.45\textwidth]{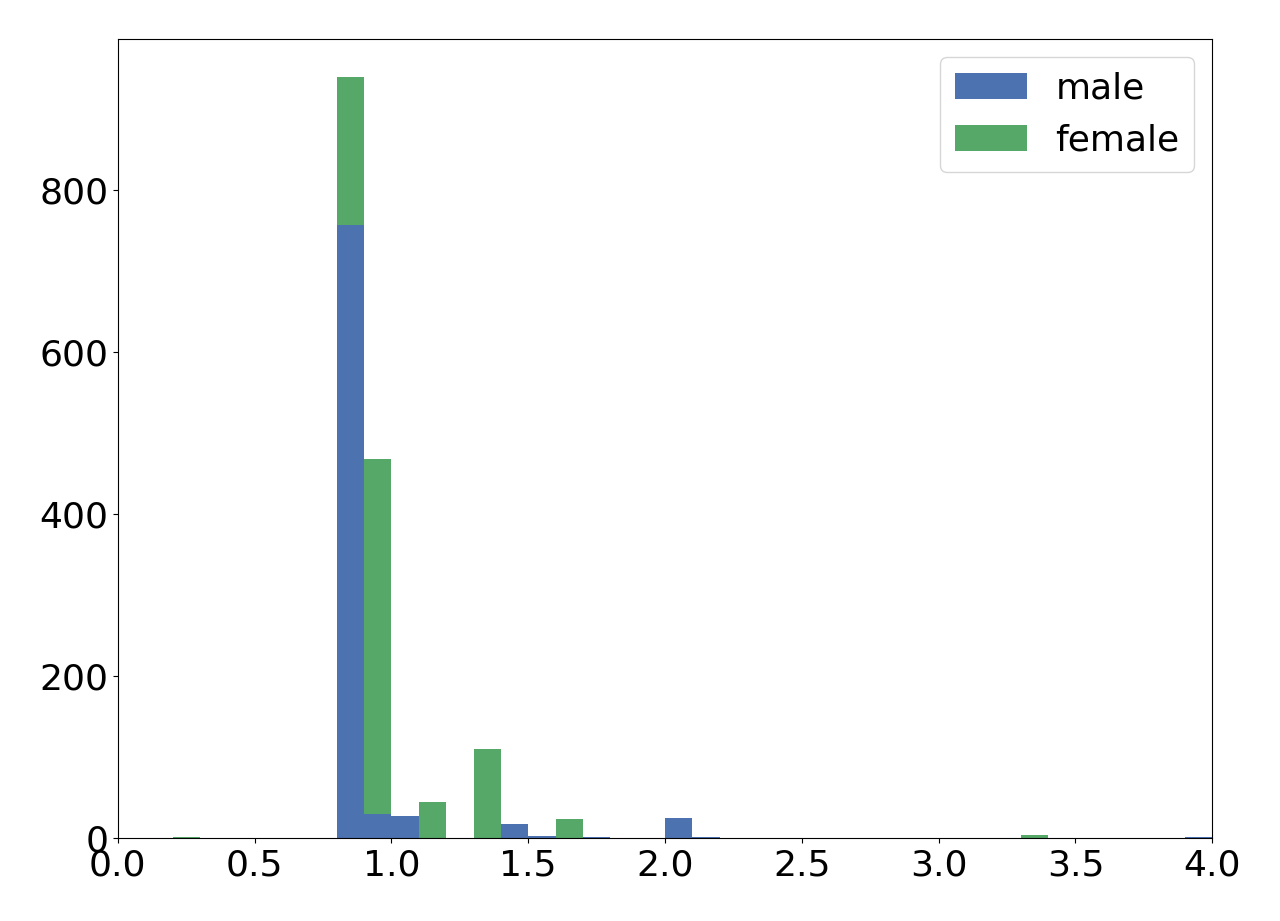}
  \caption{Histogram showing the distribution of W\textsubscript{t}-Bias weights, split into intervals of size $0.1$. 
  The blue parts of the columns correspond to masculine examples, while the green parts correspond to feminine examples. 
  The weights are centered around $1$. The largest weight ($7.68$) is not included.} 
  \label{weights-trimmed}
\end{figure}

An analysis of W-Bias weights shows that the distribution of weights contains some outliers, that is, examples with unusually large weights. 
Ten examples with largest weights have a weight average of $6.29$ (the average weight overall is $1.0$). 
These examples all come from examples in $D_k$ and $N_k$ with large $k$, because these sets are often highly gender-imbalanced, as discussed in Appendix~\ref{Appendix-distribution}.
While this is theoretically correct, it may be undesirable, as it means that few out-of-distribution examples carry a lot of weight in the final score, which could introduce noise.

A visualization of W-Bias weights of examples is shown in  Fig.~\ref{weights}.
It confirms that weights gravitate around $1$ despite the constraints.
Moreover, fewer than $1\%$ of the weights are set to $0$.
A manual investigation into these examples shows that they are often very long text spans with long lists of names, such as family trees or cast lists.
Several of them are not grammatically correct sentences, but rather lists from Wikipedia that were not removed during the annotation.

There are $9$ weights larger than $4.0$ that are not pictured:
$4.84$, $4.97$, $4.97$, $5.43$, $6.41$, $7.05$, $7.05$, $9.20$, and $9.72$, all of them corresponding to male examples.
Their weights are large, because they are coming from highly gender-imbalanced sets of examples that are highlighted and further discussed in Appendix~\ref{Appendix-distribution}.

We show that such large weights can be avoided by removing highly-imbalanced subsets of the data.
We introduce a trimmed $W$-score, called $W_t$-score.
Examples with more than $15$ personal names and examples where the correct candidate is the $k$-th closest for $k\geq 5$ are removed from this score, reducing the size of the dataset to $1670$ examples ($83.5\%$ of the original size).
Numbers $k\geq 5$ and $15$ personal names were selected manually by consulting figures that can be found in Appendix~\ref{Appendix-distribution}.
The rest of the examples are assigned new weights with the introduced method.
Top ten weights of $W_t$-score have a weight average of $3.3$, strongly reducing the problem of outliers.
Comparing $W_t$-bias with $W$-bias in Table~\ref{table-baselines-reweighted} shows that such outliers mainly affected Dist-$2$ and Dist-$3$ baselines.

A visualization of W\textsubscript{t}-Bias weights in Fig.~\ref{weights-trimmed} shows that trimming largely solves the problem of outliers, as the largest few weights now carry much less weight than before. 
Moreover, there are no examples with weight $0$.
Female weights are slightly larger on average, because there are more male ($865$) than female ($805$) examples in the trimmed dataset.
We did not conduct any additional trimming to avoid decreasing the size of the dataset further.

\subsection{Evaluation of Bias in Existing Coreference Models}

\begin{table*}[ht!]
     \begin{tabular}{@{\ }l|ccccc@{\ }}
    & $F_1$ & $F_1$-Bias & acc-Bias & W-Bias & W\textsubscript{t}-Bias \\ \hline
    \textsuperscript{1}\textsc{Bert} & $0.500$ & $0.88$ & $0.86$ & $0.85$ & $0.87$ \\ 
    \textsuperscript{1}\textsc{Bert\_WikiCREM} & $0.590$ & $0.95$ & $0.93$ & $0.90$ & $0.92$\\ 
    \textsuperscript{1}\textsc{Bert\_Gap} & $0.752$ & $0.99$ & $0.97$ &$0.96$ & $0.97$\\
    \textsuperscript{1}\textsc{Bert\_Dpr} & $0.612$ & $1.00$ & $0.96$ &$0.94$&$0.96$\\
    \textsuperscript{1}\textsc{Bert\_all} & $0.760$ & $1.03$ & $1.03$ &$1.03$ & $1.04$\\ 
    \textsuperscript{1}\textsc{Bert\_Gap\_Dpr} & $0.704$ & $1.01$ & $1.00$ & $1.00$& $1.00$ \\ 
    \textsuperscript{1}\textsc{Bert\_WikiCREM\_Gap} & $0.778$ & $1.01$ & $1.00$ & $1.00$ & $1.01$\\
    \textsuperscript{1}\textsc{Bert\_WikiCREM\_Dpr} & $0.646$ & $0.99$ & $0.98$ & $0.97$ & $0.96$\\ 
    \textsuperscript{1}\textsc{Bert\_WikiCREM\_all} & $0.783$ & $1.02$ & $1.01$ & $1.00$ & $1.01$\\ \hline
    \textsuperscript{2}\textsc{Bert\_base} & $0.824$ & $0.97$ & $0.97$ & $0.96$ & $0.96$\\ 
    \textsuperscript{2}\textsc{Bert\_large} & $0.856$ & $0.97$ & $0.96$ & $0.96$ & $0.97$\\ \hline
    \textsuperscript{3}\textsc{SpanBert\_base}  & $0.855$ & $0.96$ & $0.95$ & $0.95$&$0.95$\\ 
    \textsuperscript{3}\textsc{SpanBert\_large} & $0.877$ & $0.95$ & $0.94$ & $0.93$ & $0.93$\\ \hline
    \textsuperscript{4}\textsc{e2e}& $0.733$ & $0.93$ & $0.92$ & $0.92$ & $0.91$\\ \hline
    \textsuperscript{5}\textsc{e2e\_adv} & $0.747$ & $0.93$ & $0.91$ & $0.93$ & $0.90$\\ \hline 
    \textsuperscript{6}\textsc{RefReader} & $0.794$ & $0.96$ & $0.95$ & $0.97$ & $0.97$\\ \hline 
    \end{tabular}\ \ \ \ \ \ \ \ \ 
    \begin{minipage}{0.3\textwidth}
    \textsuperscript{1}\cite{kocijan19emnlp,kocijan2019acl};
    \textsuperscript{2}\cite{joshi2019coref};
    \textsuperscript{3}\cite{joshi2019spanbert};
    \textsuperscript{4}\cite{lee2018};
    \textsuperscript{5}\cite{SubramanianRo19}; \textsuperscript{6}\cite{RefReader}.
    We used publicly shared code and models in all cases, except for Referential Reader \cite{RefReader}, where code was not publicly available at the time. Instead, the evaluation was performed on the results provided by the authors. The numbers differ from the paper, as the authors averaged results over several seeds, but only shared one version. The results from \citet{joshi2019coref} differ from their paper, as the author shared a different checkpoint.\end{minipage}
    \caption{Evaluation of several state-of-the-art models for coreference resolution on \textsc{Gap}, with several bias scores reported.
    }
    \label{table-results}
\end{table*}

Having shown that the introduced measure strongly reduces the impact of the observed imbalances in the data, we re-evaluate recent models for coreference resolution.
Following \citet{Gap}, we consider systems that detect name spans for inference automatically and access labelled spans only to output predictions.
We thus do not consider models that were submitted at the Kaggle competition on the \textsc{Gap} dataset, because they do not conform to this norm~\cite{KaggleGap}.
The results are reported in Table~\ref{table-results}.

Comparing acc-Bias and W-Bias, we can see that only a few models change their bias score visibly, indicating that not all models were equally affected by the observed imbalances.
While we cannot directly compare these bias scores with the original $F_1$-based bias metric, we hypothesize that the imbalances in the data also affected that score.
Comparing W-Bias and W\textsubscript{t}-Bias shows that most of the models were minimally affected by the outliers in the weights.

We observe that the better performing models tend not to change their bias scores significantly.
We hypothesize that they are less affected by the observed imbalances in the data distribution.
At the same time, a larger denominator (female score) in the bias formula results in a smaller absolute difference.
Similarly, we can see that RNN-based models (models \textsuperscript{4,5,6}) change their scores more than transformer-based models (models \textsuperscript{1,2,3}), implying that RNN-based models were more affected by the number of candidates and the distance between the correct candidate and the pronoun than transformer-based models.

\paragraph*{Statistical significance of bias metrics.}
We used the randomization test \cite{StatSignificance} to compare the  W\textsubscript{t}-Bias scores of a few models, listed in Table \ref{table-results}. 
E.g., the difference between \textsc{Bert} and \textsc{Bert\_Gap} is significant ($p=0.024$), as is the difference between \textsc{Bert\_WikiCREM} and \textsc{Bert\_Wiki\-CREM\_Gap} ($p=0.017$).
Finetuning \textsc{Bert} on \textsc{Gap} thus seems to significantly increase its bias score, implying that its predictions are not as biased.
On the other hand, the difference of the \textsc{e2e} and \textsc{e2e\_adv} models is not significant ($p=0.364$), implying that the seemingly negative impact of adversarial sampling could be a coincidence.

\section{Related Work}
\label{related-work}
An increasing amount of work has recently been done on fairness both in NLP and machine learning in general.
NLP datasets for bias detection mainly detect gender bias in coreference resolution~\cite{WinoBias,WinoGender, Gap}, however, other types of bias-detection datasets exist. For example, \textsc{Eec}~\cite{SemanticAnalysisBias} is focused on sentiment analysis and additionally measures racial bias.
All listed datasets other than \textsc{Gap} are constructed artificially, often from hand-written templates. 
Thus, they do not suffer from the irregularities observed in \textsc{Gap}, however, they also do not reflect the bias on the real-world data.
We refer to~\citet{BiasSurvey} for an overview of fairness in machine learning, and to~\citet{NLPBiasSurvey} for a more specific review on biases in NLP.

Quite some work has been done in debiasing or otherwise balancing training data.
\cite{WordEmbeddingBias1,WordEmbeddingBias2,WordEmbeddingBias3} analyze and debias word embeddings, while \cite{LabelBias1,LabelBias2} focus on biased training labels.
\citet{WinoBias} show how swapping gender of pronouns and antecedents reduce the gender bias of co\-re\-fe\-ren\-ce models.
\citet{DataWeightingForClassification} and~\citet{BiasInDiscussionForums} weight training examples to remove bias from the training data, the latter also using linear programming, similarly to us. 
However, they only balance the data with respect to a single property, while our method works for several.
\citet{WinoGrande} aim to reduce the bias in coreference resolution caused by annotation artifacts both in the training and test data.
Their main aim is to remove the systemic bias in the dataset that could give away unintended cues on the correct answers.
We highlight that their work concerns a different type of bias, as their goal is to prevent any model from achieving a high performance due to spurious correlations in the dataset, rather than to reduce any type of discrimination between different candidates.

\section{Summary and Outlook}
\label{Summary}
In this work, we introduced a test-set weighting method to re\-mo\-ve undesired imbalances in 
bias-measuring da\-ta\-sets, without exacerbating other potentially undesired patterns.

We demonstrated the method on the \textsc{Gap} test set, which contained such undesired irregularities.
We annotated the dataset with spans of all personal names and introduced the bias metrics W-Bias and W\textsubscript{t}-Bias that balance out the observed irregularities.
While there is no guarantee that these scores balance out all data irregularities in \textsc{Gap}, we showed that they balance out the ones that we are aware of.
We encourage research to use our introduced metrics to measure the bias of coreference models on \textsc{Gap}.
Among the two introduced scores, we recommend W\textsubscript{t}, because its weights contain fewer outliers, and the chance of undesired deviations in the score is smaller.

A room for improvement of the method is to remove the need to identify the biases in the data, however, this step is common to existing methods that deal with bias.
It is not unreasonable to expect that the existence of bias has to be noticed and demonstrated before one can start planning the debiasing.
This already satisfies the prerequisites to use the introduced method.
Manual annotation of examples like the one in our work is not always necessary, as automatic tools (e.g.,\ NER systems) can be used.
However, 
manual annotation likely ensures the high quality of the test data.

This work addresses an important problem of real-world bias metrics. 
Future work includes balancing out unobserved irregularities in the data, extension to non-linear metrics, such as $F_1$-score, and construction of dataset that contain fewer undesired patterns that could affect the bias metric.

\section*{Acknowledgments}
The authors would like to thank Mandar Joshi and Fei Liu for their help with using their models or by sharing their predictions.
We would like to thank Ralph Abboud for his help with the proofs.
This work was supported by a JP Morgan PhD Fellowship, the Alan Turing Institute under the EPSRC grant EP/N510129/1, the AXA Research Fund, the ESRC grant ``Unlocking the Potential of AI for English Law'', and the EPSRC Studentship OUCS/EPSRC-NPIF/VK/1123106.
We~also acknowledge the use of the EPSRC-funded Tier 2 facility JADE (EP/P020275/1) and GPU computing support by Scan Computers International Ltd.
\bibliography{aaai2021}

\appendix
\newpage
\section{Proof of Theorem~\ref{upper_bound_theorem}}\label{upper_bound_theorem_proof}
We derive the criterion function for the set $A$, as the derivation for set $B$ is analogous.
We denote $$T_C := \lvert\text{Acc}^A_W(C\cap A) - \text{Acc}^A(C\cap A)\rvert.$$
We first simplify $T_C$, piece by piece:
\begin{align*}
    \text{Acc}^A(C\cap A)
        &= \frac{|C\cap A|}{|A|}
\end{align*}
\begin{align*}
    \text{Acc}^A_W(C\cap A)
        &=\frac{2}{n}\sum_{x_i\in C\cap A}w_i\,.
\end{align*}
Combining this with the formula for $T_C$, we get:
\begin{align*}
    T_C &= \frac{2}{n}\left(\sum_{x_i\in C\cap A}w_i\right) - \frac{|C\cap A|}{|A|}\\
        &= \frac{2}{n}\sum_{x_i\in C\cap A}(w_i - \frac{n}{2|A|})\,.
\end{align*}
To minimize $|T_C|$, we have to minimize $$|\sum_{x_i\in C\cap A}(w_i-\frac{n}{2|A|})|.$$

To minimize the upper bound, we take a look at the scenarios that give the largest value of $|T_C|$.
Let $w_{A(i)}$ be the $i$-th smallest weight corresponding to an example in $A$.
We use the constant $\lambda := \frac{n}{2|A|}.$
The following properties hold:
\begin{align*}
    \sum_{i=1}^{|A|}w_{A(i)} &= \sum_{x_i\in A}w_i = \frac{n}{2}\\
    \sum_{i=1}^{|A|}(w_{A(i)}-\lambda)&=\sum_{i=1}^{|A|}(w_{A(i)}-\frac{n}{2|A|}) \\
        &= \sum_{x_i\in A}w_i-\frac{n}{2} = 0\,.
\end{align*}
In the scenario where $T_C$ is maximal, $C\cap A$ will include $|C\cap A|$ examples with either largest or smallest weights.
Let $k:=|C\cap A|$, and let us first take a look at the case where examples with largest $k$ weights are in $C\cap A$.
To minimize all such cases, we aim to minimize the following term:
\begin{align}
    \sum_{k=1}^{|A|}\sum_{i=\frac{n}{2}-k+1}^{|A|}(w_{A(i)}-\lambda) = \sum_{i=1}^{|A|}(w_{A(i)}-\lambda)i \,.\label{min_term}
\end{align}
On the other hand, we aim to maximize the opposite case, i.e., when examples with the smallest $k$ weights are in $C\cap A$.
The objective that we aim to maximize can be written as follows:
\begin{align*}
    &\sum_{k=1}^{|A|}\sum_{i=1}^k(w_{A(i)}-\lambda) = \\          
        =&-\sum_{k=1}^{|A|}\sum_{i=k+1}^{|A|}(w_{A(i)}-\lambda)\\
        =&-\sum_{i=1}^{|A|}(w_{A(i)}-\lambda)i+\sum_{i=1}^{|A|}(w_{A(i)}-\lambda)\\
        =&-\sum_{i=1}^{|A|}(w_{A(i)}-\lambda)i+0\\
        =&-\sum_{i=1}^{|A|}(w_{A(i)}-\lambda)i\,.
\end{align*}
We can see that maximizing this term is equivalent to minimizing term~\eqref{min_term}.
Minimizing term~\eqref{min_term} is therefore sufficient.
We further simplify it:
\begin{align*}
    &\sum_{i=1}^{|A|}(w_{A(i)}-\lambda)i=\\
    =& \sum_{\substack{x_i,x_j\in A\\i>j}}\text{max}(w_i-\lambda,w_j-\lambda)+\sum_{x_i\in A}(w_i-\lambda)\\
        =& \sum_{\substack{x_i,x_j\in A\\i>j}}\text{max}(w_i-\lambda,w_j-\lambda)+0\\
        =& \sum_{\substack{x_i,x_j\in A\\i>j}}\text{max}(w_i,w_j)-\lambda\cdot \frac{|A|(|A|-1)}{2}\,.
\end{align*}
Since the second part of the term is constant, we aim to minimize the following objective:
\begin{align*}
        \sum_{\substack{x_i,x_j\in A\\i>j}}\text{max}(w_i,w_j)\,.
\end{align*}
In the same way, the objective for the examples in set $B$ can be computed. Summing them up, we obtain the following objective:
\begin{align*}
    \text{min}\sum_{\substack{x_i,x_j\in A\\i>j}}\text{max}(w_i,w_j)+\sum_{\substack{x_i,x_j\in B\\i>j}}\text{max}(w_i,w_j)\,.
\end{align*}

\section{Analysis of the \textsc{Gap} Data Distribution}
\label{Appendix-distribution}

\begin{figure}
  \centering
  \includegraphics[width=0.45\textwidth]{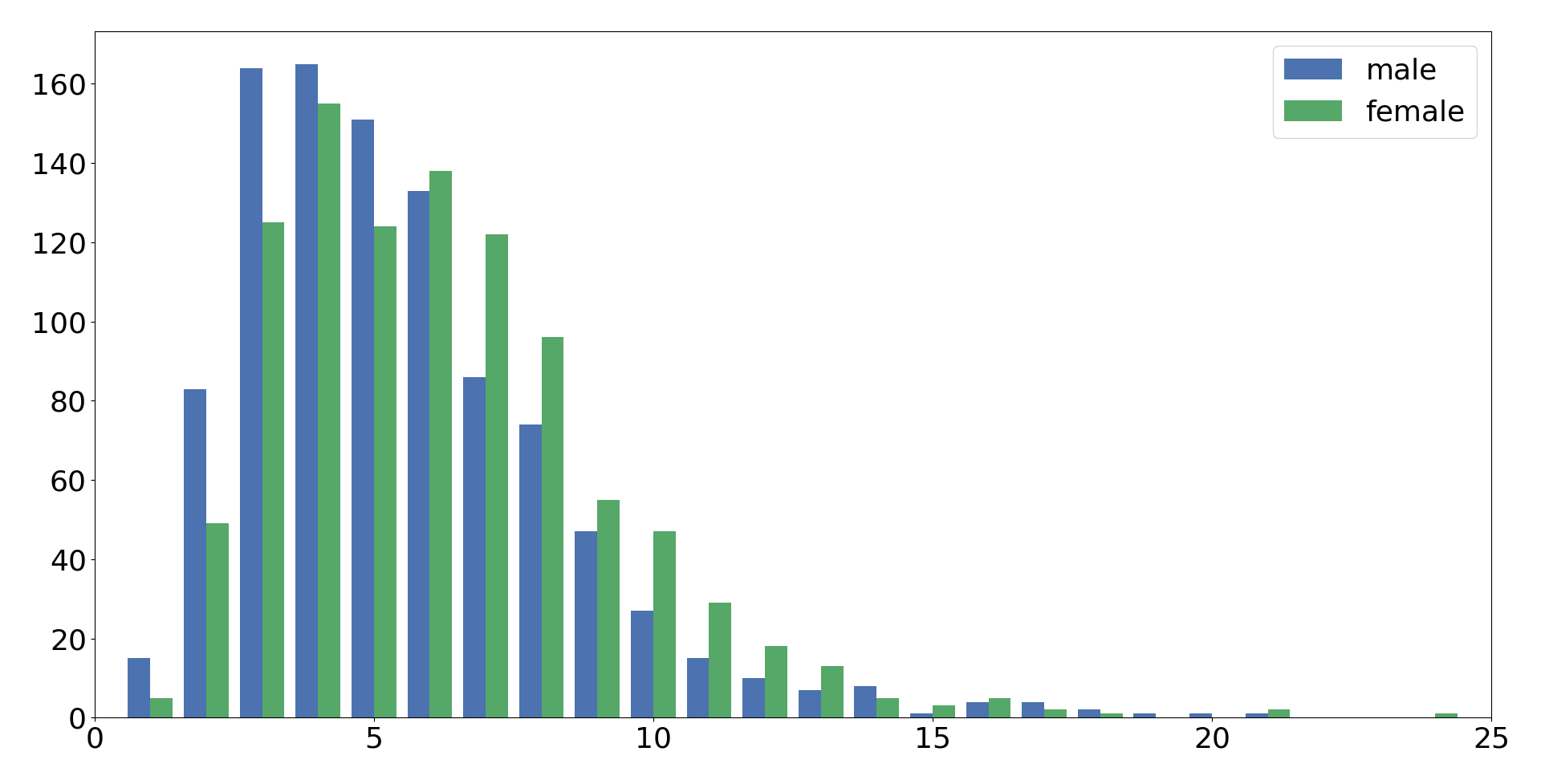}
  \caption{Histogram showing the number of personal names per sentence in the \textsc{Gap} dataset. The X-axis shows the number of names in the sentence, and the Y-axis  the number of sentences with the corresponding number of personal names. The blue and green columns show the data for masculine and feminine examples, respectively.}
  \label{n_names}
\end{figure}

This section contains information on the distribution of the \textsc{Gap} test set.
Fig.~\ref{n_names} shows the distribution of the number of names per sentence in the masculine and the feminine subset of the test set.
The difference between the masculine and the feminine subset are visible, with masculine examples usually containing fewer names per sentence in comparison to feminine examples.
This histogram clearly shows why the random baseline did not achieve a bias score around~$1$.

\begin{figure}
  \centering
  \includegraphics[width=0.45\textwidth]{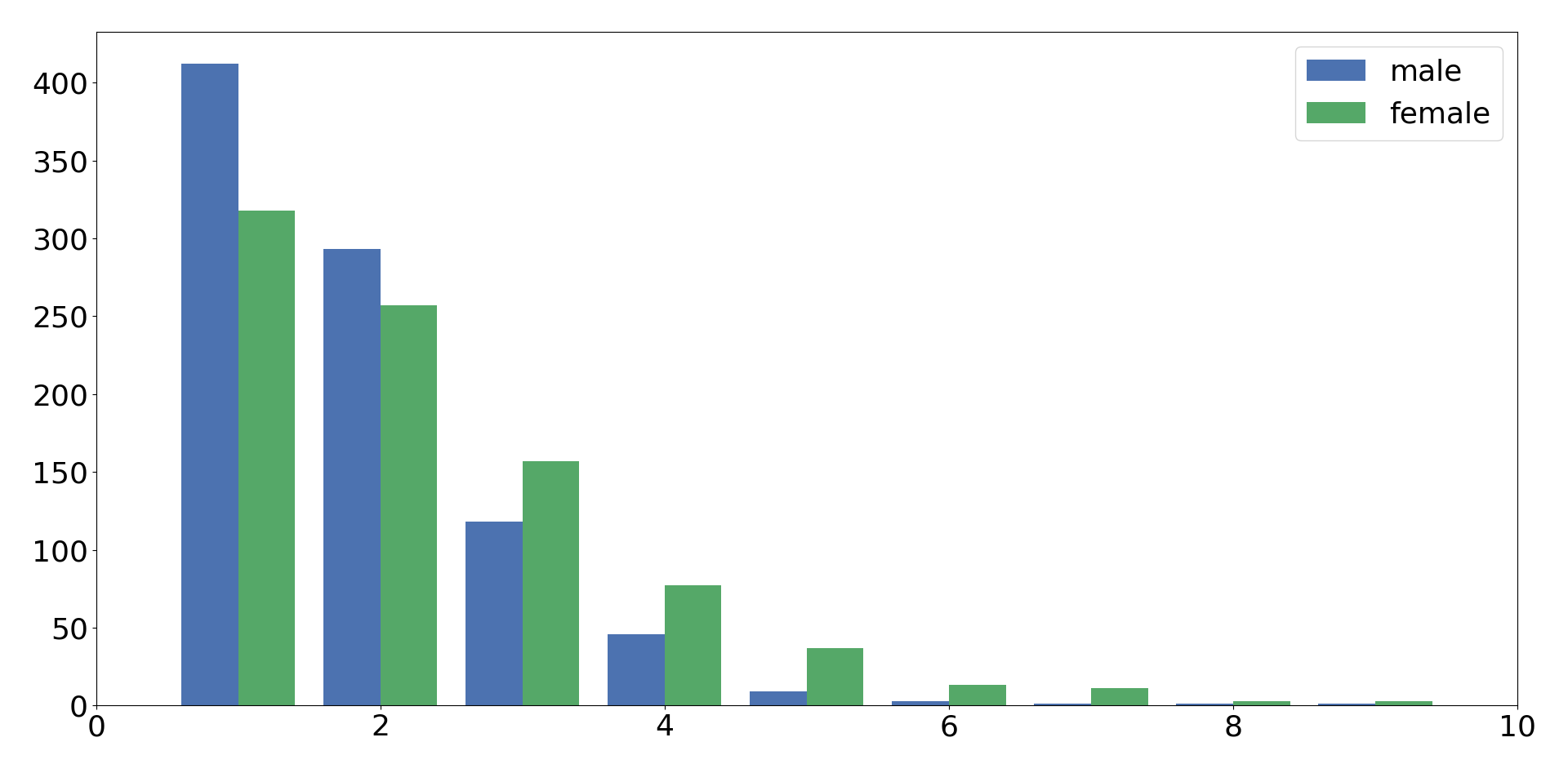}
  \caption{Histogram showing how often the correct entity is the closest, second closest, third closest, etc.\ entity to the pronoun in the \textsc{Gap} dataset. The blue and green columns show the data for masculine and feminine examples, respectively.}
  \label{distance}
\end{figure}

In Fig.~\ref{distance}, we can see how often the correct entity is the closest one to the pronoun, how often it is the second closest, third closest, etc.
We can observe a similar pattern, with the closest and second closest candidate being correct more often in the masculine subset.
The distribution is the source of a seemingly biased performance of the Dist-$k$ baselines.

We can see that sets at the tails of both two graphs can be highly gender-imbalanced.
As discussed in Sections~\ref{section-weighting} and~\ref{weights_appendix}, the weighting method counteracts these imbalances by assigning large weights to the weights in the under-re\-pre\-sen\-ted class.

\begin{figure}
  \centering
  \includegraphics[width=0.45\textwidth]{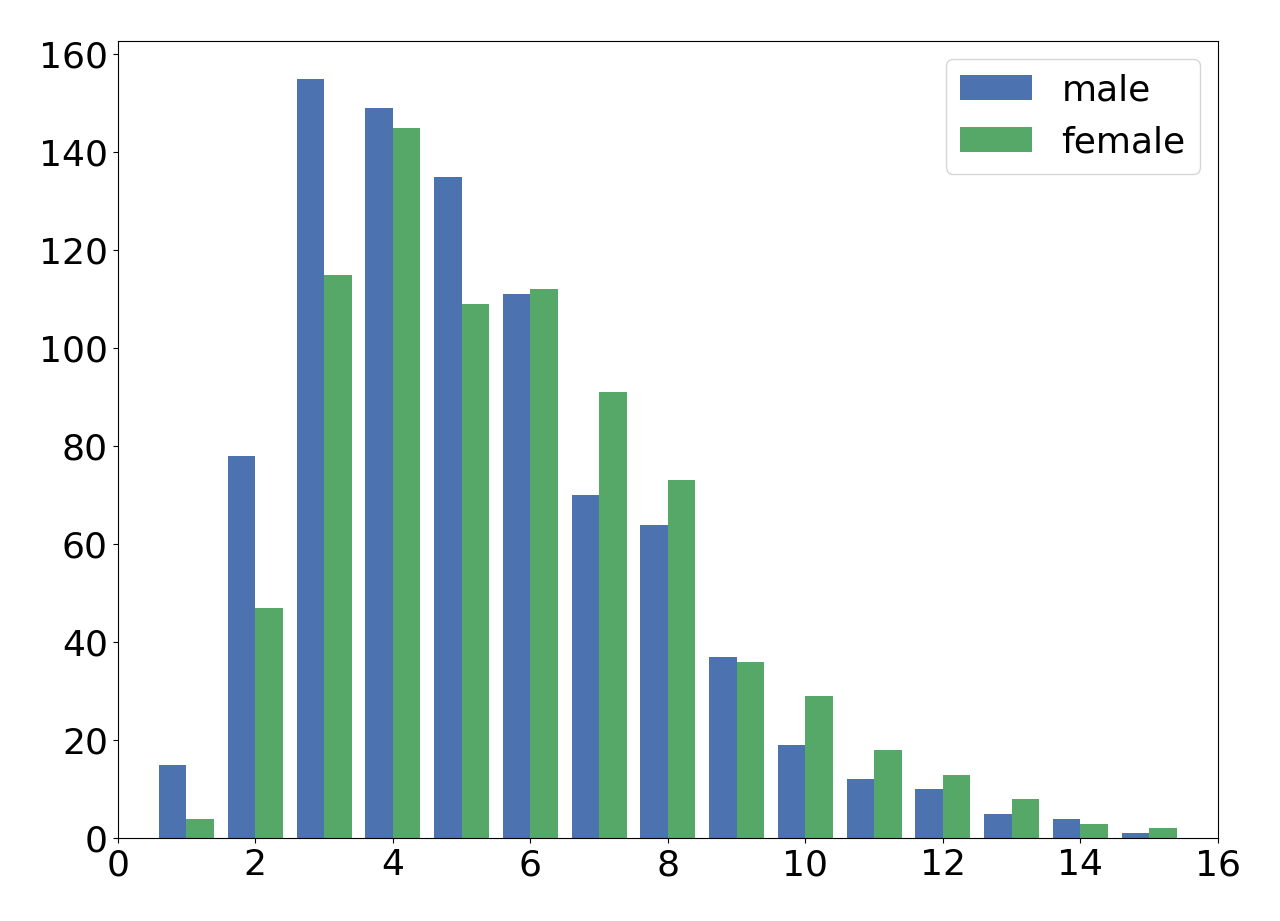}
  \caption{Histogram showing the number of personal names per sentence in the \textsc{Gap} dataset after the trimming conducted for the W\textsubscript{t}-Bias score. The blue and green columns show the data for masculine and feminine examples, respectively.}
  \label{n_names_after_trimming}
\end{figure}

\begin{figure}
  \centering
  \includegraphics[width=0.45\textwidth]{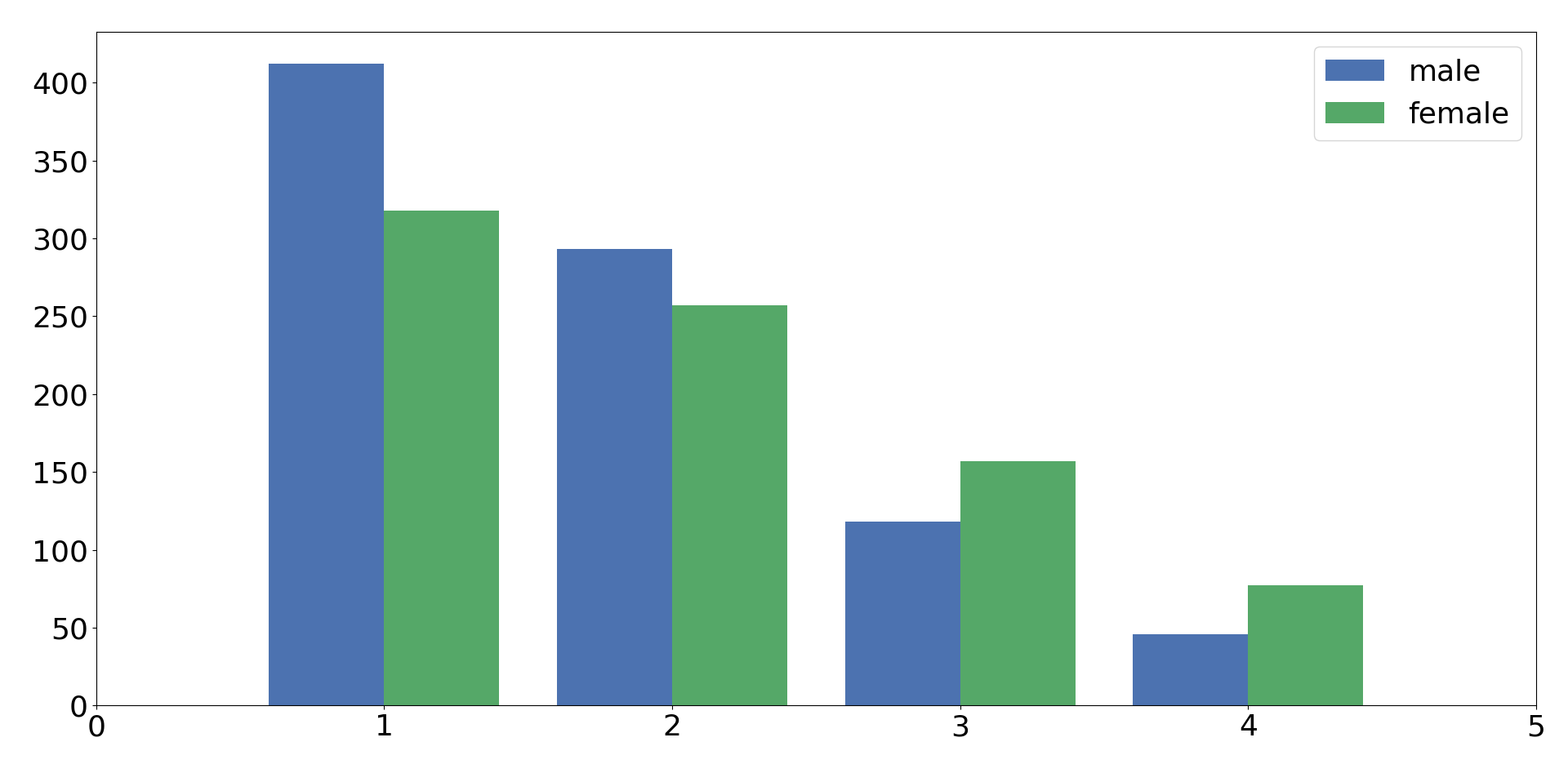}
  \caption{Histogram showing how often the correct entity is the closest, second closest, third closest, etc.\ entity to the pronoun in the \textsc{Gap} dataset after the trimming conducted for the W\textsubscript{t}-Bias score. The blue and green columns show the data for masculine and feminine examples, respectively.}
  \label{distance_after_trimming}
\end{figure}

We additionally visualize the dataset after the trimming that was done as part of the W\textsubscript{t}-Bias score.
The distribution of the number of names and how close the correct candidate is after \textit{trimming} are reported in Figs.~\ref{n_names_after_trimming} and~\ref{distance_after_trimming}.
The trimmed dataset contains fewer highly-imbalanced subsets.
We did not conduct any further trimming of the dataset to avoid a further reduction of its size.

\end{document}